\documentclass{article}
\usepackage{ijcai22}

\usepackage{natbib}
\usepackage{times}
\usepackage{soul}
\usepackage{url}
\usepackage[hidelinks]{hyperref}
\usepackage[utf8]{inputenc}
\usepackage[small]{caption}
\usepackage{graphicx}
\usepackage{amsmath}
\usepackage{amsthm}
\usepackage{booktabs}
\usepackage{algorithm}
\usepackage{algorithmic}
\usepackage{xcolor}

\title{What AI can do for horse-racing?}

\author{
    Pierre Colle\and
    Alezan AI\\
    \emails
    pierre@alezan.ai
}

\begin{document}

\maketitle

\begin{abstract}
Since the 1980s, machine learning has been widely used for horse-racing predictions, gradually expanding to where algorithms are now playing a huge role as price makers in the betting market.

Machine learning has changed the horse-racing betting market over the last ten years, but main changes are still to come. The paradigm shift of neural networks (deep learning) may not only improve our ability to simply predict the outcome of a race, but it will also certainly shake our entire way of thinking about horse-racing - and maybe more generally about horses.

Since 2012, deep learning provided more and more state-of-the-art results in computer vision and now statistical learning or game theory. Following \cite{Tuyls2020}, we describe how the convergence of the three machine learning fields (computer vision, statistical learning, and game theory) will be game-changers in the next decade in our ability to predict and understand horse-racing.

We consider that horse-racing is a real world laboratory where we can work on the animal-human interaction and build a non-anthropocentric Artificial Intelligence. We believe that this will lead us to understand the horses better and the interactions between animals and humans in general.
\end{abstract}

\section{Introduction}

The current work has been largely inspired by the recent work regarding AI in football (\cite{Tuyls2020}). This article's structure will follow the same structure by first studying the different fields of AI and then the frontiers between those fields.

We will study AI from the viewpoint three different axes:

\begin{itemize}
\item  Computer Vision \textbf{(CV)} -- AI analyzes videos and images to extract information

\item  Statistical learning  \textbf{(SL)} -- build models and predict the race outcome.

\item  Game theory \textbf{(GT)} -- modelling cross-competitor interaction and understanding their strategies against each other.
\end{itemize}

Within these axes, we will go further into each frontier :

\begin{itemize}
\item  Frontier 1: (\textbf{SL and GT}) -- how to combine statistical learning with game theory

\item  Frontier 2: (\textbf{CV and SL}) -- how to combine Computer Vision with statistical learning

\item  Frontier 3: (\textbf{CV and GT}) -- how to combine computer vision with game theory
\end{itemize}

Several works have been published regarding soccer, baseball (\cite{kelly-0000}), and basketball (\cite{li-2021}) in the last few years, but much less has been published regarding horse-racing. This is mainly because of the betting-centric nature of the horse-racing ecosystem that makes it very secret. Anyone who has a good idea about horse-racing will keep it to himself and not share his betting advantage with other punters, as secrecy is often the best strategy to protect intellectual assets (see \cite{cohen2000protecting}). This is even more important in parimutuel betting, where punters are not betting against the casino, but against one another.


\begin{figure*}[t]
\centering
\includegraphics[width=8in, height=5in,keepaspectratio=true]{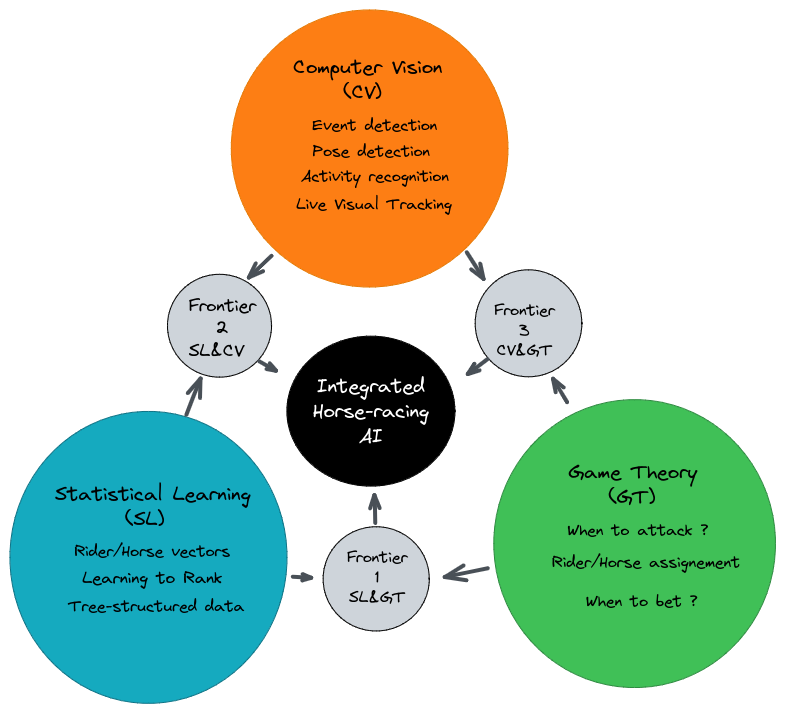}
\caption{Diagram of the different fields of AI and the corresponding Frontiers, inspired by \cite{Tuyls2020}}
\label{fig:ai-horse-racing}
\end{figure*}

\section{AI Fields}

\subsection{Statistical Learning}

Since the 19th century, horse-racing has been a special field -- maybe the most iconic -- for statistical analysis. In the late 20th century with the emergence of computer capabilities, statistical learning has been used to predict the outcome of races and then to ``beat the odds''.

By 2018, the algorithms used for betting had deeply changed the horse-racing industry (\cite{malachane-2018}). This market is dominated by a few syndicates containing many workers -- up to 300 employees (according to \cite{Ziemba2020}). Those syndicates are not publishing or communicating their work. 

Current work is limited to public knowledge, not discussing the actual methods used by betting syndicates.

The first usage of statistical learning is to build a ``handicapping model''  (see \cite{Benter-1994}), which basically predicts a score on each horse about his ability to win the race. To build such a model, one can leverage a large amount of data available for each horse. This usually includes:

\begin{itemize}
\item  The horse's past results

\begin{itemize}
\item  the amount of money earned

\item  lifetime records
\end{itemize}

\item Information about the breed, like the dosage index (\cite{roman-2003}).
\end{itemize}

We are also gradually switching to ``big-data'' approaches based on larger datasets like:

\begin{itemize}
\item  Genomic data (see \cite{danvy-2018})

\item  GPS tracking, Nano-bio sensors (see \cite{evans-2020})

\item  Horserace videos and images, as discussed in Section \ref{computer-vision}
\end{itemize}

The main challenge for statistical learning in horse-racing is to leverage a large amount of data, with different structures inside an integrated end-to-end trainable system.

These heterogeneous big data situations will raise more challenges for feature engineering (see \cite{wang2017heterogeneous}). The perspective of using deep learning to automatically focus on key data seems to be an obvious yet challenging opportunity.

\subsubsection{Handling Historical Data}

The most important information about horses on a track is their past records. 

\begin{figure}
\includegraphics[width=3.5in, height=2.4in, keepaspectratio=true]{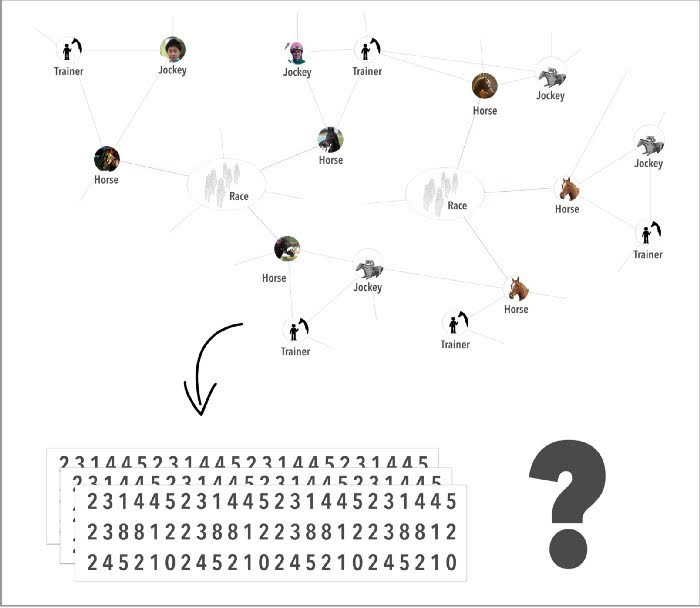}
\caption{Image reproduced from \cite{deep-learning-and-horse-race-prediction}
with the authorisation of the author, Ko Ohashi}
\label{fig:graph}
\end{figure}

Past records are time-series, and the ability to manage time series is still a challenging topic (see \cite{Deep-learning-2019}). However, when looking more closely at the past records of a horse, we would like to consider not only as a time series but more as a tree-structured graph (see figure \ref{fig:graph} from \cite{deep-learning-and-horse-race-prediction}).


The state-of-the-art of the deep-learning on tree-structured data is still moving fast, and while some works are using RNN-based architecture (see \cite{Tree2Tree-18,dong-2016,Tree-structured-17}), the complexity and sparsity of the tree structures make the resulting network rather slow (see \cite{Harer2019TreeTransformerAT})


In the context of horse-racing, to train an end-to-end network understanding historical data, we will face two major issues:
\begin{itemize}
\item Sparsity -- most horses participate in 15 to 30 races in their lives (see \cite{deep-learning-and-horse-race-prediction}). There are about 10,000 horses trained for thoroughbred racing in France (see \cite{chiffres-cles-france-galop}), so most of the horses never face one another.

\item Complexity -- given that each race has 10 competitors, to end-to-end train a network while considering the historical performance of the 10 last races for each horse, there will be about 100 000 competitors (input vectors) to take into account.

\end{itemize}

\subsubsection{Building the Loss}

The other main challenge for an efficient end-to-end trained handicapping model is to build the loss.

The naive approach would be to work on a classifier winner/not winner, and then use a binary cross-entropy loss (see example in \cite{deep-learning-and-horse-race-prediction}). 

This classifier may give first results but will only use a tiny portion of the available data, considering that the first horse is the only interesting thing to take into account while training the neural network. More complete usage of data will make it possible to leverage all the available data :

\begin{itemize}
\item Final rank information
\item Distance on arrival between horses
\item Detailed race proceedings
\end{itemize}

To use the final rank information and the distance on arrival between horses, the learning-to-rank (see \cite{liu-2007}) information-retrieval research seemed to tackle this exact problem, and libraries are now available publicly to work on this (see for example \cite{pasumarthi-2019}), the main idea here is to consider that what matters in not only who is the winner but what is the exact ordering of the output.

The detailed proceeding of the race is also difficult to leverage in a loss function, but it could be possible to efficiently leverage the data by using counterfactual analysis as described in Frontier 1 (see section \ref{frontier1}).

\subsection{Computer Vision}\label{computer-vision}

Today's Computer Vision systems can tackle horse-racing and numerous challenges with human-based sports like:
\begin{itemize}
\item  Activity recognition

\item  Pose estimation -- studying the way the horse is moving and if he suffers from specific injuries. (\cite{UAD-Animal-Pose})


\item  Tracking the horses in the race (see for example \cite{augment-betting-experience})

\end{itemize}

These technologies work well in labs, but it is still challenging to connect them to real-world broadcasting systems and to pass the usability test for real-world applications.

\begin{figure}
\includegraphics[width=3.2in, height=4.5in, keepaspectratio=true]{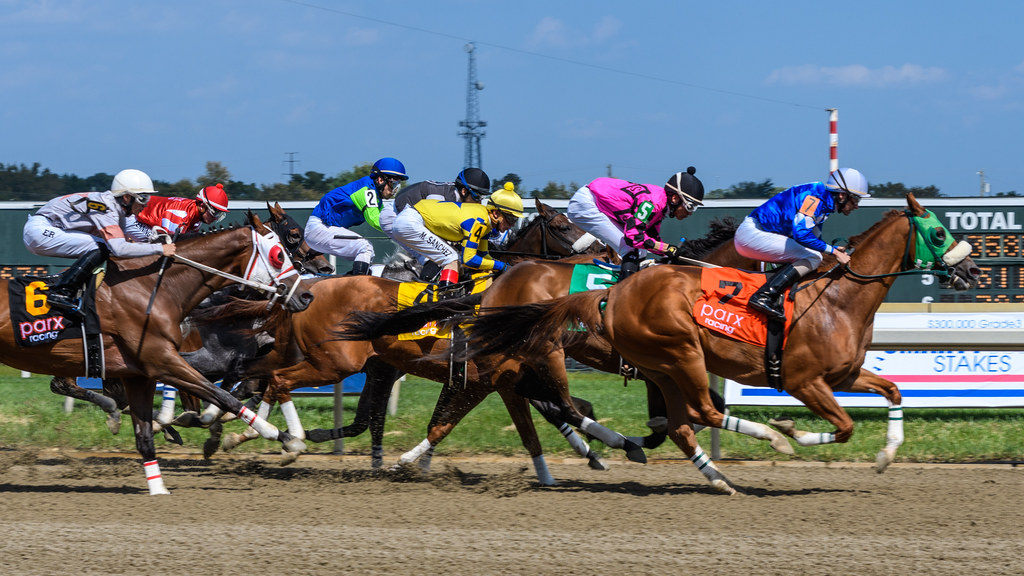}
\caption{Horses often occlude one another in a horse race and it is challenging to precisely draw bounding boxes around each one. Photo made by Peter Miller, from https://www.flickr.com/photos/pmillera4/40952321920}
\label{fig:occlusion}
\end{figure}

The main issues with real-world data are:
\begin{itemize}
\item The constant changes of cameras in broadcasting systems, and the need to geometrically combine them (as described for football in \cite{Tuyls2020})
\item The number of real-life scenarios (see a famous example in \cite{the-verge})


\item The high number of occlusions present in horse-racing real-life data, where all the horses are running side by side (see Figure \ref{fig:occlusion})
\end{itemize}

A lot more usages for computer vision will arise in subsequent years like emotion recognition or real-time data extraction, but these usages will often be related also to Game Theory or Statistical Learning, those are discussed in Frontier 2 (see section \ref{frontier2}) and Frontier 3 (see section \ref{frontier3}).

\subsection{Game Theory} \label{gt}

Game theory is about players behavior and decisions. In horse-racing with strategic decisions are made at different levels :

\begin{itemize}
\item Season level: Participating or not in a race
\item Race level: What is the strategy ?
\item Action level: When to draft and when to attack ?
\end{itemize}


\begin{figure}[ht!]
\includegraphics[width=3.30in, keepaspectratio=true]{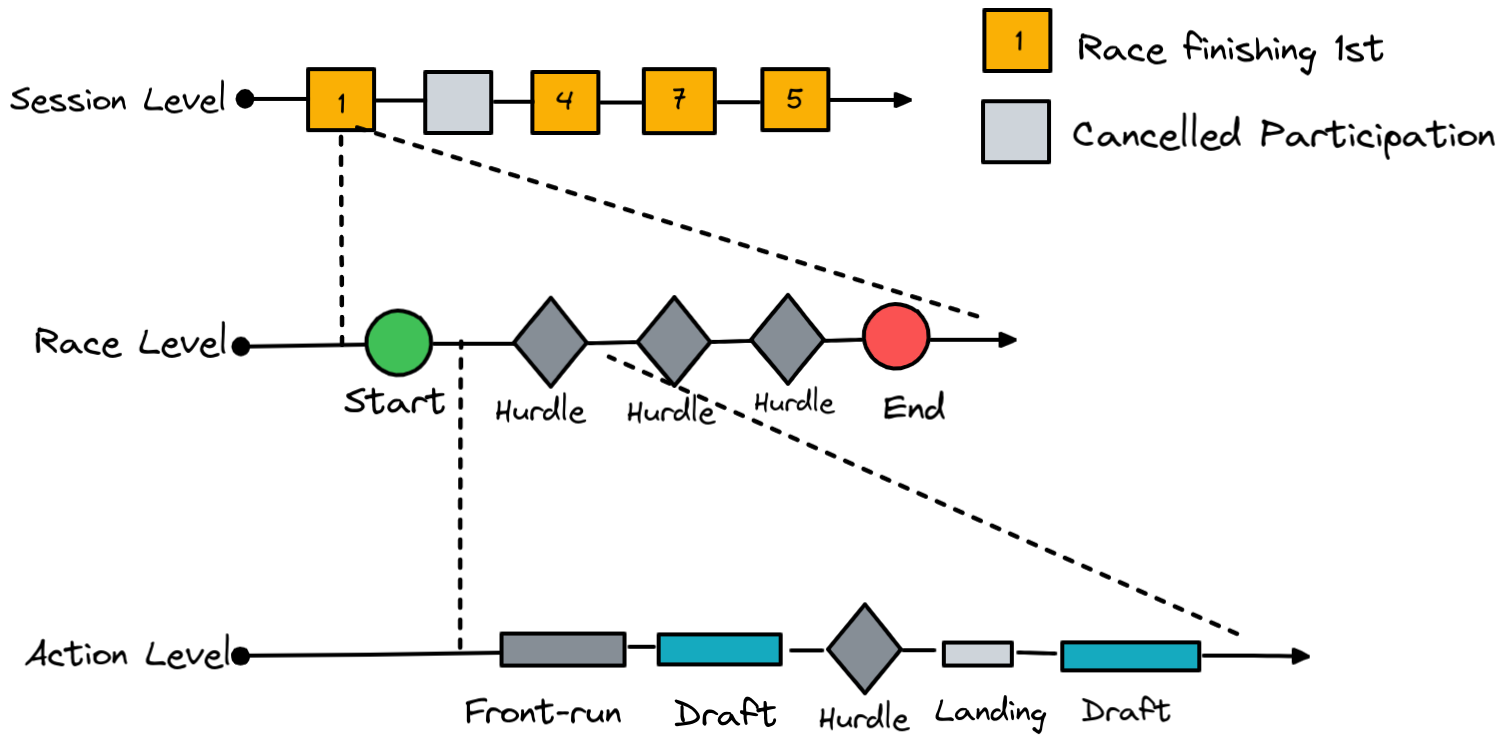}
\caption{}
\label{fig:gametheory}
\end{figure}

\subsubsection{Season Level: Participating or not in a race}

When an owner makes one of his horses participate in a race, the choice is made for a reason, and this decision can be viewed as a game-related decision. From the owner's point of view, he can get the following benefits:
\begin{itemize}
\item  The allocation (money paid to the winner, but also the 2nd, 3rd, 4th and 5th horse)

\item  The visibility of the horse may raise its value on the horse market
\end{itemize}

But each race also has a cost for the owner:

\begin{itemize}
\item  The horse can be injured and lose its value on the market

\item  Participating in a race costs money, food, transportation
\end{itemize}

We could ask for each competitor at the beginning of the race:

\begin{itemize}
\item  Why is this horse participating in this race?

\item  What is his target rank (in the owner's point of view)?
\end{itemize}

Apart from the participation of the race, the owner (and the trainer) needs to decide under what conditions the horse will participate to this race:
\begin{itemize}
\item  Which rider should run with the horse? This may influence not only the short-term outcome of the race but also the long-term welfare of the horse (\cite{MUNSTERS201275} have linked, for show-jumping the welfare of the horse with the choice of the rider).

\item What type of blinker (Winkers/blinkers/none) ? Depending on the horse's character, blinkers are a way to limit his field of view and to limit intimidation by the crowd/other horses.
\item What type of horseshoe (Barefoot/Horsehoes) ? While Barefoot are faster, this is also more dangerous for the horse health
\item Training intensity and warm up intensity and duration
\end{itemize}

\subsubsection{Race Level: What is the Strategy?}

Specific parts of the race may be amenable to game theory, like the beginning of the race, the positioning inside or outside the race field, or deciding how to balance the effort over the race. The first strategical dilemma is related to the type of race. Is it a slow, medium, or fast race?

It has been studied for other sports like running. By studying the medallists in the 1500m Championships, it is clear that the kick of the race (the final sprint) in the last lap depends on how fast the previous laps were (\cite{sandford-2019}). The winning strategy regarding when to start the sprint in a race depends on the speed of the race.

In any case, this ``kick'' approach of the 1500m running races is not as relevant for 400m or 800m distances, as the world-best records are made with athletes who can run the first half of the race faster than the second half. This suggests that for his best performance, an athlete needs to reach a high-intensity effort during the whole race that will impact his performance in the final meters (\cite{sandford-2019}).

The regulation has been studied for horse-racing also, and it's seems that it is more similar to the 400m or 800m as the winning horse is often the one that is able to slow down the least at the end
(see \cite{mercier-2020}).

These considerations regarding ``kick'' and strategy are very important for horse-racing because unlike running where the muscle and the strategic brain belong the same individual, the horse-rider who decides the tactical aspects of the race needs to manage the muscles of a horse. This muscle/brain duality makes horse-racing a very interesting topic to study for AI (see section \ref{frontier1})

\subsubsection{Action Level: When to Draft and When to Attack}

In the race, each rider needs to handle his tactic between attacking or drafting. This is well explained in the cycling-related literature (see \cite{mignot-2016}), but it can be easily extended to horse racing as the average speed of horses was found to increase with the percentage of the race spent on a drafting position (\cite{spence-2012}).

The question of whether to use whipping or how to use it is also a scientific research topic, and had been discussed regarding horse-welfare. The amount of whip-related pain for horses whereas recent studies have shown that whipping has no impact on horses' speed (see \cite{mcgreevy-2012} and \cite{wilson-2018}), and the question of whether whipping is an aid for the steering (and for the game-theoretic aspect of the horse riding) is still a matter of debate in the community (\cite{thompson-2020}).
\section{AI Frontiers}

\subsection{Frontier 1: Interaction Between Horse and Rider (GT and SL)}\label{frontier1}

The combination of Game Theory and Statistical Learning is a main topic for horse-racing handicapping models. One of the key improvement in next years AI (as discussed for football in \cite{Tuyls2020}, and implemented for basketball in \cite{baller2vec}) is to train reinforcement learning to predict the horse's next move and use this to compare the predicted moves with the existing ones.


We may consider that statistical learning will give us the ability to train, using reinforcement learning, the typical behavior of this specific horse in a typical situation. Then, by comparing the expectations from statistical learning with the actual behavior of the horse  (counterfactual analysis), we could extract the ``unexpected'' behaviors, which in the context of machine learning could be interpreted as  ``tactical'' moves.

Traditionally, for analysis purpose, horse races are divided by predefined sectors, first sector are the 400 first meters, last sector are the 400 last meters. This split gives the ability to consider the race sector per sector and to get information not only about the arrival, but also about the proceeding of the race. 

Considering the large amount of information given by big data approach, the first benefit of counterfactual analysis - (as explained in \cite{Tuyls2020} - is to train an attention mechanism given the ability, user counter-factual analysis, to focus on relevant sequences. Then we would replace a non-trained handmade sector traditional point of view, by a dynamic, trainable attention-based mechanism.

The main idea of counter-factual analysis is to compare the actual behavior of a horse with the expected behavior of the horse in our model. This difference could be, following the mathematical definition of information by Shanon (see \cite{shannon1948mathematical}), seen simply as an information extraction, or a compression mechanism, focusing only on the unexpected.

Another benefit, By assuming that the tactical moves are made by the rider and by doing the statistical learning on the horse only (with horse historical data), one may even use this counterfactual analysis to extract the rider's strategic moves (see \cite{spence-2012}]) from the (less strategical) horse behavior.

This analytical split between horse-related and rider-related components will be very useful to assess the role of the horse and the rider on a specific result or ranking, and then it can provide statistical learning with some very accurate analysis to work on.




\subsection{Frontier 2: SL and CV}\label{frontier2}

Computer Vision can increase the amount of data available for statistical learning. A lot of important data are available right now (GPS trajectories, historical data), but some other information is still only in the raw videos. High dimensional videos and the ability to extract that information on a large scale might feed new statistical learning models.

Some computer-vision accessible data are:
\begin{itemize}
\item  An event such as horsewhips and falls (In \cite{whip-fall}, researchers are using video recording as a primary data source to study the horsewhips)

\item  The position of the horses' legs and the frequency of their movement (As in \cite{mathis2021pretraining})
\item  The horses' emotions (nervousness, attention level). Some studies show that the horses' facial expressions, like the orientation of the head and position of the ears and eyes, yield information about the horses' attention (see \cite{eyes-and-ears-are-visual-indicators}).

\item The horse's eagerness to enter the starting gates. This information is already scrutinized by professional punters. (see Educating the Punter [1994])


\end{itemize}

The first goal for this kind of study would be, obviously, to improve handicapping model; but this deep image-to handicapping approach can be extremely informative to understand the horse's (and more generally animals) behavior.
The main advantage of deep learning compared to other approaches is the ability to remove the bias in the study and then let the neural network decide what to look at in a specific image.

As explained in \cite{montavon2018} we are able to use visualization techniques to understand meaningful patterns in the input data and understand, which part of an image plays a role in final neural network decision.

Having the ability to train an end to end neural network from videos to race prediction, may give a first answer to the question of whether the ears' position before the race has been used by the network to predict the outcome of the race. This new approach will may anyway gives new tools to study, on a large scale basis, the correlations between horse emotions, visual markers and race result.


\subsection{Frontier 3: GT and CV}\label{frontier3}

In the same way that Computer Vision can be used to improve statistical learning on Frontier 2, we can also use it to feed game-theoretic models. A lot of Computer-Vision accessible information can provide relevant information for game-theoretic models such as:

\begin{itemize}
 \item A warm-up session format: knowing how long and how fast the warm up is will give us information about the horse and the owner expectations.
 \item The global appearance of the horse and some very simple considerations like the horse's hair dress will give us an idea if the horse owner expects to show his horse in this race.
 \item The body movements of the horse-rider will provide information about his relationship with the horse.
\end{itemize}

\begin{figure}
\includegraphics[width=3.2in, height=4.5in, keepaspectratio=true]{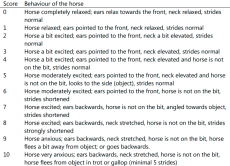}
\caption{Behavioral score used to assess the nervosity of a horse in \cite{MUNSTERS201275}. Reproduced with the authorization of the author}
\label{fig:nervousness}
\end{figure}

A study regarding horse welfare (\cite{MUNSTERS201275}) used a behavioral score built mainly on visual observations (see Figure \ref{fig:nervousness}). It shows that this behavioral score can be used to measure the matching between a horse and its rider. Only 21 riders and 16 horses were included in the study

Using Computer Vision and working on horse racing, we will be able to generalize this study for every broadcasted race; instead of working on 16 horses, we could extract this kind of behavioral score (see Figure \ref{fig:nervousness}) for thousands of horses and riders competing each year in a league.

Changing the order of the magnitude of this data extraction will create a lot of opportunities. First of all, it may impart the ability to characterize horse and rider personality, by creating a rider-temperament vector for each rider and a horse-temperament vector for each horse, as it is made in soccer for players in ?.

Once we have a rider-temperament vector for each rider and a horse-temperament vector for each horse, it would be quite direct to give an ``impairment score'' to every horse-rider couple even if they had never ridden together before. This may be used to help the coaches choose the rider to ride a specific horse.

But this vector approach once trained using the numerous available videos of horse racing could be used outside of horse-racing for other equine usages: to match a horse with a rider in contexts like show-jumping when less data are available but the behavioral score and horse-rider matching could be transferred.

Eventually, as discussed in section \ref{frontier2}, working with a huge horse-racing video dataset gives the opportunity to rebuild a behavioral score analysis and let the neural network decide exactly what to look for in the image to decide whether the horse and the rider are matching or not.
It sounds feasible that AI will teach us (humans) how and what to look and horses, without having all the anthropocentric bias we tend to project on horses when we look at them.

\section{Conclusion: What AI can do for horse-racing}

In the next decades, AI may have the ability to fulfill some completely new scenarios on horse-racing. These include:
\begin{itemize}
\item Improving the odds and making them much more relevant, while taking more and more information into consideration.
\item Augmented reality will improve the way people watch horse-racing, making it more fun to watch horseraces and bet on a specific horse.
\item Splitting the performance between the strategic moves of the rider and the physical performance of the horse.
\item Making an ``impairment score'' between a horse and a rider to help horse trainers in their choices.
\item Telling the trainer, owners, and the public exactly what to look at before and during the race to best predict a horse's ability.
\item Teaching how to look at horses, how to understand their behavior, when they are nervous, and where they are not.
\end{itemize}

Our final concern is to know whether this AI revolution will be made in public labs and published research, aiming to spread technologies, tools, and knowledge to everyone, or if it will be made privately and secrecy with only short-term interest.

In the last decades, the recent AI revolution (with the example of Facebook and Google) needed both public and open-source research communities working together along with private interests. We feel that AI will encourage horse-racing and the betting ecosystem to open themselves, open-source  algorithms, and share papers and knowledge. 

\section{Discussion: What Can horse-racing do for AI?}

Horse racing is an open laboratory with a huge amount of existing data, existing private funding, and real-life competition, but it is still missing a public research communities and widespread technologies for Computer Vision.

Current AI is a very anthropocentric field with, for example, human-only pose estimation and human-only emotion recognition. It would be interesting for AI to focus on a less anthropocentric approach and use horse racing as a large-scale real-world laboratory to experiment with the ability of AI to predict human-animal-related outcomes.

Finally, we always ask ourselves what we can teach AI, but maybe we should think more about what AI can teach us? The ability for AI in terms of horse racing to help us understand the horse's emotion is an opportunity for it to create knowledge about an animal's emotions and characters.

\nocite{mayton-2014}
\nocite{pasumarthi-2019}
\nocite{fereira-2018}

\bibliographystyle{model5-names}
\bibliography{ijcai22}

\end{document}